\documentclass[runningheads]{llncs}
\usepackage[T1]{fontenc}
\usepackage{graphicx,verbatim}
\usepackage{booktabs}
\usepackage{graphicx}
\usepackage{multirow}
\usepackage{makecell}
\usepackage{amsmath}
\usepackage[nolist]{acronym}
\usepackage{color}
\usepackage{hyperref}

\begin{document}
\title{Beyond Classification: Pathology Foundation Models as Detection Encoders for Mitotic Figures}

\titlerunning{Foundation Model Features as Detection Encoders for Mitotic Figures}
%

\newif\ifpeerreview

\ifpeerreview
\author{Anonymized Authors}  
\authorrunning{Anonymized Author et al.}
\institute{Anonymized Affiliations \\
    \email{email@anonymized.com}}

\else

\author{Sweta Banerjee\inst{1,6} \and
Alireza Teimoury\inst{1} \and
Nils Porsche\inst{1}\and
Alexandra K. Stoll\inst{6,7}\and
Viktoria Weiss\inst{2}\and
Niklas~Hargarter\inst{1}\and
Jonas Ammeling\inst{4}\and
Thomas Conrad\inst{3}\and
Christoph Stroblberger\inst{5}\and
Christopher Kaltnecker\inst{5}\and
Robert Klopfleisch\inst{3}\and
Christof~A.~Bertram\inst{2}\and
Katharina Breininger\inst{6}\and
Marc Aubreville\inst{1}}

\authorrunning{Banerjee et al.}
\institute{Flensburg University of Applied Sciences, Flensburg, Germany \and
 University of Veterinary Medicine, Vienna, Austria \and
 Freie Universit\"{a}t Berlin, Berlin,  Germany \and
 Technische Hochschule Ingolstadt, Ingolstadt, Germany \and
 Medical University of Vienna, Vienna, Austria \and
 Julius-Maximilians-Universit\"{a}t W\"{u}rzburg, W\"{u}rzburg, Germany \and 
 Friedrich-Alexander-Universit\"{a}t Erlangen-N\"{u}rnberg, Germany
}

\fi
\maketitle              
\begin{abstract}
Pathology foundation models (FMs) are models trained on vast amounts of typically unlabeled data and have been shown to yield regularized latent spaces that can be used effectively in downstream classification tasks. This is also true for the classification of mitotic figures vs.\ other cells. However, it is so far unclear if the latent space of current FMs provides features that are discriminant and spatially suitably resolved to also serve as a backbone for dense object detection paradigms. 
In this work, we investigate this question for common current pathology FMs (UNI, UNI2-h, Virchow, Virchow2, H-optimus-0, H-optimus-1) and compare their performance against a fully end-to-end trained baseline based on a ResNet50 architecture. We combine FM backbones with representatives of single stage, dual stage and self-attention-based detectors (RetinaNet, Faster R-CNN, Deformable DETR respectively) on the multi-domain MIDOG++ dataset, and on the TUPAC16 dataset as an out-of-domain case. 
We show that the H-optimus-0 and Virchow models yielded competitive performance, indicating that the latent spaces of current FMs---all trained on image-level self-supervision---are suitable for direct mitotic figure detection and may be slightly more robust on our out-of-domain test case. All code is made available publicly at \url{https://anonymous.4open.science/r/FM4MFdet-24CA/README.md}.

\keywords{Foundation Models  \and Mitotic Figures \and Object Detection.}

\end{abstract}

\section{Introduction}

Mitosis is the process by which a cell divides into two daughter cells, enabling the proliferation that supports tissue growth, renewal, and the replacement of senescent cells. A dividing cell, when observed under a microscope, presents as a \ac{MF}. Beyond their physiological role in normal tissue maintenance, MFs serve as an important prognostic marker across a range of human and veterinary cancers, where uncontrolled cell division is a hallmark of malignancy~\cite{azzola2003tumor,bertram2024mitotic}. In these tumors, the density of \acp{MF}, typically quantified as the mitotic count, guides treatment decision-making~\cite{fitzgibbons2023protocol,mcniel1997evaluation}.
Despite its prognostic importance, manual \ac{MF} assessment by pathologists suffers from limited reproducibility and is time-consuming in practice. Mitotic counting first requires identifying the hotspot within an entire tissue, which is constrained by the limited time available for tissue screening, and second to identify all mitotic figures within that hotspot, which is hindered by variable decision thresholds among pathologists~\cite{bertram2022computer}. 

Recently, pathology \acp{FM} such as UNI, H-Optimus, and Virchow have emerged as powerful, general-purpose feature extractors, demonstrating strong performance across a broad array of histopathology benchmarks \cite{campanella2025clinical,neidlinger2025benchmarking}. They were trained on large-scale, unlabeled datasets by means of self-supervised learning using schemes such as DINO~\cite{caron2021emerging,oquab2023dinov2} or MoCo~\cite{he2020momentum,chen2021empirical}. All these schemes have in common that they use a similarity-based scoring mechanism that disentangles augmented images from the same origin from those of different origins. This paradigm regularizes the latent space to yield discriminatory features for downstream tasks. However, the overwhelming majority of these evaluations was carried out solely on patch- or image-level classification tasks \cite{ammeling_benchmarking_2026,melba:2026:006:banerjee}, and a few on segmentation tasks \cite{chen2026extent}. A separate line of work integrates FMs into cell-level detection and segmentation pipelines, such as CellViT \cite{horst2024cellvit}, CellViT++ \cite{horst2026cellvit++}, and CellSAM \cite{marks2025cellsam}, which couple FM encoders with dedicated decoders for nucleus instance prediction. Notably, when CellViT++ was applied to \ac{MF} detection on MIDOG++, it fell short of a trained-from-scratch RetinaNet detector across organs~\cite{horst2026cellvit++}, illustrating that strong FM-based cell representations do not necessarily transfer to the demands of mitosis detection. It is thus so far unclear whether strong performance on classification benchmarks translates to utility on dense-prediction tasks such as object detection, which demand fine-grained spatial localization rather than image- or patch-level semantic summarization. 

In recent works, the detection of \acp{MF} has largely relied on architectures based on ImageNet-pretrained network stems that are fully fine-tuned end-to-end on the detection task~\cite{aubreville2024domain,aubreville2026mitosis}. Pathology FMs were not yet evaluated on this task. Whether the frozen latent representations of pathology FMs encode sufficient information to compete with such detectors trained end-to-end with all parameters updated is thus an open question with significant practical implications, particularly since studies evaluating FMs for \ac{MF} classification indicate that the \ac{OOD} performance of FM-based models can potentially exceed that of fully trained networks, given the well-regularized training objective~\cite{ammeling_benchmarking_2026,melba:2026:006:banerjee}; we therefore also investigate this behavior in the object detection setting.

In this work, we therefore ask: do the frozen latent spaces of pathology \acp{FM} encode sufficiently dense information to detect \acp{MF}, compared to a convolutional baseline whose backbone is fine-tuned on the task? We address this through a systematic evaluation of frozen FM backbones against an ImageNet-pretrained ResNet-50 detector trained end-to-end on \ac{MF} detection.

\section{Materials and Methods}

\begin{figure}[t]
    \centering
    \includegraphics[width=\textwidth]{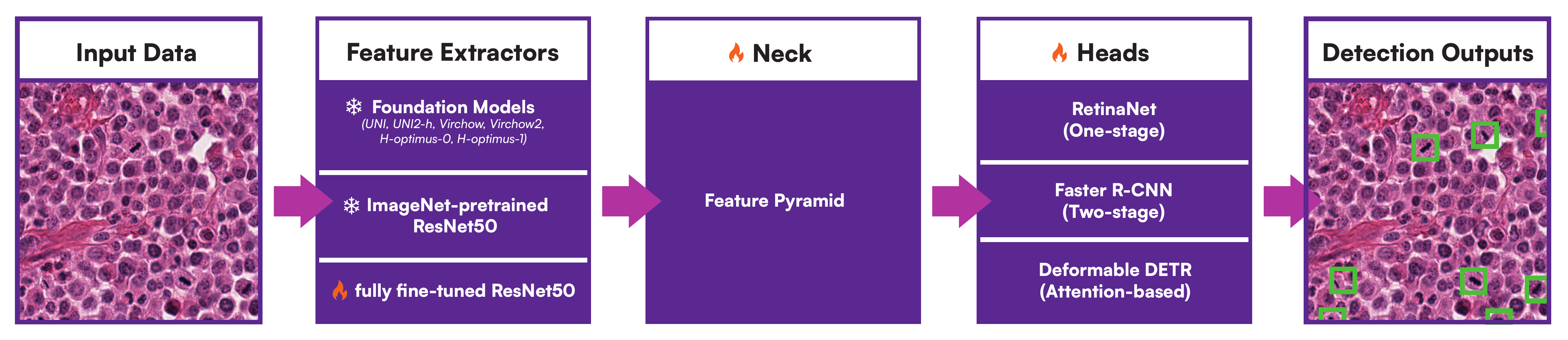}
    \caption{Overview of the detection pipeline.}
    \label{fig:pipeline}
\end{figure}

We compare frozen pathology \acp{FM} with a conventional convolutional baseline on mitotic figure detection (Fig.~\ref{fig:pipeline}). We evaluate six ViT-based FMs -- UNI~\cite{chen2024towards}, UNI2-h~\cite{chen2024towards}, Virchow~\cite{vorontsov2024virchow}, Virchow2~\cite{zimmermann2024virchow2}, H-Optimus-0~\cite{hoptimus0}, and H-Optimus-1~\cite{hoptimus1} -- as frozen backbones, each combined with three detector heads (Faster R-CNN~\cite{ren2016faster}, RetinaNet~\cite{lin2017focal}, and Deformable DETR~\cite{zhu2020deformable}), yielding 18 frozen FM configurations. As baselines, we use an ImageNet-pretrained ResNet-50 in two regimes, fine-tuned and frozen, each crossed with the same three heads, for 24 configurations in total. All configurations shared the same training and evaluation splits, detailed below.

\subsection{Dataset and preprocessing}

We conducted all experiments on MIDOG++, a multi-domain \ac{MF} detection dataset comprising seven distinct domains~\cite{aubreville2023comprehensive}. The images of the dataset each represent regions of interest with a size of $2mm^2$. Images were split into training, validation, and test sets on a per-patient basis to prevent data leakage, resulting in 333/71/71 slides, respectively. For \ac{OOD} evaluation, we used the relabeled~\cite{bertram2020pathologist} training set of the TUPAC16 challenge~\cite{veta2019predicting} as secondary test set. The dataset consists of 71 individual tumor cases of human breast cancer~\cite{veta2019predicting}.
Regions of interest were tiled into non-overlapping patches chosen to align with the \ac{ViT} patch grid of each backbone. We used a resolution of 1008 and 1024 px for patch-14 and patch-16 model configurations, respectively. 
During training, tiles were sampled to including at least one \ac{MF} per patch and we used a sliding-window scheme over the full regions of interest during test-time inference.

\subsection{Frozen foundation-model backbones}

Each FM backbone is a \ac{ViT} pre-trained in a self-supervised fashion on hundreds of millions to billions of tiles~\cite{chen2024towards,vorontsov2024foundation,zimmermann2024virchow2,hoptimus0,hoptimus1}. During our experiments, these backbones were used strictly as frozen feature extractors, ensuring that we probed the representations as learned during pretraining, without any task-specific adaptation.
A \ac{ViT} encodes each image as a sequence of patch embeddings plus non-spatial summary tokens (a class token and, where applicable, register tokens). For dense detection, we discarded all non-spatial tokens and reshaped the remaining patch embeddings into a two-dimensional feature map at the token-grid resolution (one spatial location per patch token), which serves as the input to the neck. Each backbone was normalized using the same statistics employed during its pretraining: DINOv2-based models (UNI, UNI2-h, Virchow, Virchow2) use ImageNet mean and standard deviation, while H-Optimus-0 and H-Optimus-1 use their published pathology-specific normalization. We selected these six backbones since they include widely used public pathology \acp{FM} spanning different pretraining strategies and model scales, providing a representative sample of current practice.

\subsection{Feature pyramid neck}

Because ViT backbones operate at a single spatial resolution, whereas modern detectors expect a multi-scale hierarchy~\cite{lin2017focal,zhu2020deformable}, we inserted a lightweight feature pyramid neck following the ViTDet~\cite{li2022exploring} design. Starting from the single-scale ViT feature map, the neck progressively refines features towards finer resolutions and aggregates them towards coarser resolutions, constructing a multi-level feature pyramid.
For patch-14 backbones, this yields a five-level hierarchy for Faster R-CNN and RetinaNet and a four-level hierarchy for Deformable DETR, with effective strides of $[7, 14, 28, 56, 112]$ pixels relative to the input. These scales range from a high-resolution level suited for individual MFs to coarse levels providing broader contextual cues. Apart from the detection head, the neck is the only learnable module in the frozen-backbone setting; thus, it bears the full burden of adapting a fixed single-scale representation to multi-scale dense prediction. Because the neck natively outputs feature maps with the expected spatial resolutions and channel dimensions, it also replaces the intermediate channel projection that Deformable DETR would otherwise require. The ResNet-50 baseline does not use this ViTDet neck. Because ResNet already produces features at multiple scales, each detector uses its standard neck instead: a feature pyramid network over the C3--C5 stages for Faster R-CNN and RetinaNet, and a channel mapper for Deformable DETR. The baseline also keeps each detector's default strides rather than the patch-14 strides used for the ViT backbones.

\subsection{Detection heads}

We evaluated three standard detectors with fixed hyperparameters chosen from their reference implementations and literature, kept identical across all backbones within a given head. These heads span the three dominant detection paradigms---two-stage (Faster R-CNN~\cite{ren2016faster}), one-stage (RetinaNet~\cite{lin2017focal}), and transformer-based set prediction (Deformable DETR~\cite{zhu2020deformable})---so that our conclusions are not tied to a single detector family.

\emph{Faster R-CNN} used a region proposal network with anchor scale 8 and aspect ratios $\{0.5, 1.0, 2.0\}$; its RoI head comprised two 1024-channel fully connected layers with $7 \times 7$ RoI pooling over the finest four pyramid levels, with RPN IoU thresholds of 0.7/0.3 and detection thresholds of 0.5/0.5. %
\emph{RetinaNet} used anchors with an octave base scale of 4, three scales per octave, and the same aspect ratios, trained with focal loss ($\gamma = 2.0$, $\alpha = 0.25$) and L1 box regression. %
\emph{Deformable DETR} adopted the standard multi-scale deformable attention architecture with default encoder/decoder depth and query count, using Hungarian matching with focal, L1, and generalized IoU costs as in the original formulation~\cite{zhu2020deformable}. For Faster R-CNN and RetinaNet, inference used a score threshold of 0.05 and NMS at IoU 0.5.

\subsection{Training protocol}

Unless otherwise noted, all backbone-head combinations shared the same optimization and augmentation settings. All experiments were implemented using the MMDetection framework~\cite{mmdetection}, with \ac{FM} weights loaded via \texttt{timm} \cite{rw2019timm} and Hugging Face model hubs. All experiments use a fixed global random seed for reproducibility.

\paragraph{Optimization.}

We trained all configurations with frozen backbones (the frozen FMs and the frozen ResNet-50 baseline) using AdamW with learning rate $2 \times 10^{-4}$, weight decay $1 \times 10^{-4}$, and momentum parameters $\beta = (0.9, 0.999)$. Weight decay was not applied to normalization or bias parameters, and gradient norms were clipped at 35 to stabilize training. Training began with a linear learning-rate warm-up by a factor of $10^{-3}$ for the first 500 iterations. After warm-up, Faster R-CNN and RetinaNet used a cosine learning-rate schedule decaying to a minimum of $10^{-7}$, while Deformable DETR followed its standard step schedule with a single $10\times$ drop.
All models were trained for up to 100 epochs with validation-based early stopping: training terminated if the validation mAP failed to improve by at least $10^{-3}$ for 10 consecutive epochs (Faster R-CNN and RetinaNet) or 20 epochs (Deformable DETR). We used a batch size of 16 for training for all configurations.

\paragraph{Data augmentation.}

We applied random horizontal and vertical flips (probability 0.5), small random affine transformations (rotation up to $\pm 15^\circ$, translation up to $\pm 5\%$ of image size, scaling between 0.9 and 1.1), HED stain jittering, and photometric distortion.



\paragraph{Evaluation} 
We calculated the micro-averaged F1 (aggregated over all images), as well as precision and recall, as is commonly done for \ac{MF} detection benchmarking~\cite{aubreville2023comprehensive,veta2019predicting} as thresholded metrics. Moreover, we calculated the area under the free-range operating characteristic curve (FROC), interpolated over up to eight false positives per image as done in the most recent MIDOG 2025 challenge~\cite{aubreville2026mitosis}.
\section{Results}

\begin{table}[t]
\centering
\caption{Detection performance across frozen foundation-model backbones,
detection heads, and ResNet-50 baselines (trained end-to-end and frozen) on the
MIDOG++ test set. All metrics are reported with 95\% bootstrap confidence
intervals (10{,}000 slide-level resamples) in brackets; F1, precision, and
recall are evaluated at each configuration's validation-selected operating
threshold, and FROC-AUC is a threshold-independent area on the $[0,8]$ FP/image
scale. The best value in each column is shown in \textbf{bold}.}
\label{tab:midogpp_results}
\setlength{\tabcolsep}{4pt}
\renewcommand{\arraystretch}{1.05}
\resizebox{\columnwidth}{!}{%
\begin{tabular}{llcccc}
\toprule
Backbone & Head & F1 & Precision & Recall & FROC \\
\midrule
UNI & Faster R-CNN & 0.4513 {\tiny [0.3851, 0.5071]} & 0.5007 {\tiny [0.3998, 0.5904]} & 0.4107 {\tiny [0.3465, 0.4749]} & 2.158 {\tiny [1.798, 2.550]} \\
UNI & RetinaNet & 0.6614 {\tiny [0.6096, 0.6992]} & 0.7556 {\tiny [0.6897, 0.8030]} & 0.5881 {\tiny [0.5329, 0.6343]} & 4.310 {\tiny [3.958, 4.663]} \\
UNI & Deformable DETR & 0.7062 {\tiny [0.6601, 0.7376]} & 0.7513 {\tiny [0.6938, 0.7985]} & 0.6663 {\tiny [0.6161, 0.7055]} & 4.751 {\tiny [4.467, 5.033]} \\
\addlinespace
UNI2-h & Faster R-CNN & 0.7559 {\tiny [0.7206, 0.7858]} & 0.7727 {\tiny [0.7207, 0.8161]} & 0.7398 {\tiny [0.6932, 0.7795]} & 4.999 {\tiny [4.515, 5.490]} \\
UNI2-h & RetinaNet & 0.7178 {\tiny [0.6801, 0.7481]} & 0.7680 {\tiny [0.7109, 0.8144]} & 0.6739 {\tiny [0.6192, 0.7276]} & 4.633 {\tiny [4.111, 5.158]} \\
UNI2-h & Deformable DETR & 0.7316 {\tiny [0.6978, 0.7579]} & 0.7962 {\tiny [0.7414, 0.8372]} & 0.6768 {\tiny [0.6403, 0.7133]} & 4.924 {\tiny [4.559, 5.302]} \\
\addlinespace
Virchow & Faster R-CNN & 0.7294 {\tiny [0.6888, 0.7649]} & 0.7885 {\tiny [0.7364, 0.8280]} & 0.6785 {\tiny [0.6292, 0.7246]} & 4.871 {\tiny [4.356, 5.373]} \\
Virchow & RetinaNet & 0.7622 {\tiny [0.7241, 0.7922]} & 0.7853 {\tiny [0.7322, 0.8270]} & 0.7404 {\tiny [0.6978, 0.7755]} & 5.252 {\tiny [4.826, 5.697]} \\
Virchow & Deformable DETR & 0.7198 {\tiny [0.6853, 0.7472]} & 0.7756 {\tiny [0.7177, 0.8193]} & 0.6715 {\tiny [0.6345, 0.7095]} & 4.708 {\tiny [4.302, 5.141]} \\
\addlinespace
Virchow2 & Faster R-CNN & 0.7545 {\tiny [0.7206, 0.7814]} & 0.7841 {\tiny [0.7406, 0.8197]} & 0.7270 {\tiny [0.6855, 0.7618]} & 5.081 {\tiny [4.640, 5.518]} \\
Virchow2 & RetinaNet & 0.7503 {\tiny [0.7200, 0.7758]} & 0.7932 {\tiny [0.7480, 0.8318]} & 0.7118 {\tiny [0.6751, 0.7473]} & 5.135 {\tiny [4.755, 5.536]} \\
Virchow2 & Deformable DETR & 0.7169 {\tiny [0.6816, 0.7499]} & 0.7767 {\tiny [0.7223, 0.8191]} & 0.6657 {\tiny [0.6215, 0.7138]} & 4.769 {\tiny [4.348, 5.237]} \\
\addlinespace
H-optimus-0 & Faster R-CNN & 0.7542 {\tiny [0.7251, 0.7789]} & 0.7990 {\tiny [0.7564, 0.8343]} & 0.7141 {\tiny [0.6754, 0.7518]} & 5.133 {\tiny [4.719, 5.547]} \\
H-optimus-0 & RetinaNet & 0.7718 {\tiny [0.7368, 0.8002]} & \textbf{0.8197} {\tiny [0.7768, 0.8542]} & 0.7293 {\tiny [0.6893, 0.7646]} & 5.490 {\tiny [5.099, 5.880]} \\
H-optimus-0 & Deformable DETR & 0.7350 {\tiny [0.6991, 0.7632]} & 0.7532 {\tiny [0.6912, 0.7995]} & 0.7176 {\tiny [0.6834, 0.7545]} & 4.829 {\tiny [4.397, 5.297]} \\
\addlinespace
H-optimus-1 & Faster R-CNN & 0.7717 {\tiny [0.7333, 0.8024]} & 0.7580 {\tiny [0.7070, 0.7983]} & \textbf{0.7859} {\tiny [0.7451, 0.8208]} & 5.171 {\tiny [4.724, 5.616]} \\
H-optimus-1 & RetinaNet & 0.7683 {\tiny [0.7263, 0.8024]} & 0.7609 {\tiny [0.7084, 0.8027]} & 0.7760 {\tiny [0.7313, 0.8140]} & 5.264 {\tiny [4.845, 5.686]} \\
H-optimus-1 & Deformable DETR & 0.6637 {\tiny [0.6276, 0.6994]} & 0.7419 {\tiny [0.6812, 0.7879]} & 0.6004 {\tiny [0.5510, 0.6619]} & 4.063 {\tiny [3.599, 4.599]} \\
\addlinespace
ResNet-50 & Faster R-CNN & 0.7702 {\tiny [0.7363, 0.7950]} & 0.7950 {\tiny [0.7504, 0.8340]} & 0.7468 {\tiny [0.7045, 0.7822]} & 5.292 {\tiny [4.954, 5.644]} \\
ResNet-50 & RetinaNet & \textbf{0.7917} {\tiny [0.7567, 0.8163]} & 0.8080 {\tiny [0.7651, 0.8443]} & 0.7760 {\tiny [0.7279, 0.8125]} & \textbf{5.686} {\tiny [5.340, 6.031]} \\
ResNet-50 & Deformable DETR & 0.7452 {\tiny [0.7124, 0.7706]} & 0.7771 {\tiny [0.7242, 0.8174]} & 0.7159 {\tiny [0.6832, 0.7472]} & 5.153 {\tiny [4.860, 5.485]} \\
\addlinespace
ResNet-50 (frozen) & Faster R-CNN & 0.7183 {\tiny [0.6768, 0.7472]} & 0.7434 {\tiny [0.6875, 0.7942]} & 0.6949 {\tiny [0.6366, 0.7426]} & 4.740 {\tiny [4.390, 5.133]} \\
ResNet-50 (frozen) & RetinaNet & 0.7166 {\tiny [0.6766, 0.7448]} & 0.7889 {\tiny [0.7301, 0.8412]} & 0.6564 {\tiny [0.6015, 0.7090]} & 4.880 {\tiny [4.559, 5.243]} \\
ResNet-50 (frozen) & Deformable DETR & 0.7498 {\tiny [0.7126, 0.7776]} & 0.7957 {\tiny [0.7500, 0.8320]} & 0.7089 {\tiny [0.6664, 0.7435]} & 5.147 {\tiny [4.763, 5.545]} \\
\bottomrule
\end{tabular}%
}
\end{table}

\begin{table}[t]
\centering
\caption{Out-of-domain detection performance, 
evaluated on TUPAC16. F1, precision, and recall are reported at each
configuration's operating threshold, with 95\% bootstrap confidence intervals
(10{,}000 slide-level resamples) in brackets; FROC-AUC was evaluated on the $[0,8]$ FP/image scale. Best value in each column is shown in \textbf{bold}.}
\label{tab:tupac_results}
\setlength{\tabcolsep}{4pt}
\renewcommand{\arraystretch}{1.05}
\resizebox{\columnwidth}{!}{%
\begin{tabular}{llcccc}
\toprule
Backbone & Head & F1 & Precision & Recall & FROC \\
\midrule
UNI & Faster R-CNN & 0.3952 {\tiny [0.3041, 0.4669]} & 0.3329 {\tiny [0.2428, 0.4120]} & 0.4862 {\tiny [0.3712, 0.5751]} & 1.773 {\tiny [1.354, 2.235]} \\
UNI & RetinaNet & 0.5381 {\tiny [0.4609, 0.5921]} & 0.6556 {\tiny [0.5664, 0.7207]} & 0.4562 {\tiny [0.3696, 0.5281]} & 2.926 {\tiny [2.434, 3.496]} \\
UNI & Deformable DETR & 0.6282 {\tiny [0.5869, 0.6598]} & 0.7014 {\tiny [0.6431, 0.7430]} & 0.5688 {\tiny [0.5130, 0.6287]} & 3.715 {\tiny [3.316, 4.236]} \\
\addlinespace
UNI2-h & Faster R-CNN & 0.7043 {\tiny [0.6645, 0.7362]} & 0.6880 {\tiny [0.6434, 0.7480]} & 0.7214 {\tiny [0.6308, 0.7818]} & 4.017 {\tiny [3.203, 4.891]} \\
UNI2-h & RetinaNet & 0.6932 {\tiny [0.6596, 0.7241]} & 0.6617 {\tiny [0.6220, 0.7153]} & 0.7279 {\tiny [0.6587, 0.7771]} & 4.195 {\tiny [3.490, 4.994]} \\
UNI2-h & Deformable DETR & 0.7188 {\tiny [0.6869, 0.7474]} & 0.7202 {\tiny [0.6816, 0.7635]} & 0.7174 {\tiny [0.6689, 0.7508]} & 4.524 {\tiny [3.908, 5.244]} \\
\addlinespace
Virchow & Faster R-CNN & 0.7054 {\tiny [0.6706, 0.7313]} & \textbf{0.7401} {\tiny [0.7036, 0.7730]} & 0.6738 {\tiny [0.6259, 0.7130]} & 4.296 {\tiny [3.675, 4.986]} \\
Virchow & RetinaNet & 0.7262 {\tiny [0.6964, 0.7499]} & 0.7131 {\tiny [0.6789, 0.7493]} & 0.7399 {\tiny [0.6917, 0.7762]} & 4.607 {\tiny [3.985, 5.308]} \\
Virchow & Deformable DETR & 0.7057 {\tiny [0.6718, 0.7300]} & 0.7278 {\tiny [0.6858, 0.7594]} & 0.6848 {\tiny [0.6374, 0.7232]} & 4.372 {\tiny [3.774, 5.012]} \\
\addlinespace
Virchow2 & Faster R-CNN & 0.6937 {\tiny [0.6567, 0.7232]} & 0.6570 {\tiny [0.6137, 0.7013]} & 0.7349 {\tiny [0.6716, 0.7808]} & 4.122 {\tiny [3.367, 4.906]} \\
Virchow2 & RetinaNet & 0.6984 {\tiny [0.6660, 0.7260]} & 0.6495 {\tiny [0.6072, 0.6982]} & 0.7554 {\tiny [0.6976, 0.7961]} & 4.437 {\tiny [3.796, 5.177]} \\
Virchow2 & Deformable DETR & 0.6824 {\tiny [0.6470, 0.7090]} & 0.6347 {\tiny [0.5896, 0.6773]} & 0.7379 {\tiny [0.6865, 0.7749]} & 4.012 {\tiny [3.303, 4.787]} \\
\addlinespace
H-optimus-0 & Faster R-CNN & 0.7168 {\tiny [0.6847, 0.7447]} & 0.7254 {\tiny [0.6835, 0.7783]} & 0.7084 {\tiny [0.6475, 0.7525]} & 4.053 {\tiny [3.103, 5.067]} \\
H-optimus-0 & RetinaNet & \textbf{0.7349} {\tiny [0.7077, 0.7585]} & 0.7182 {\tiny [0.6858, 0.7633]} & 0.7524 {\tiny [0.7011, 0.7840]} & 4.589 {\tiny [3.897, 5.407]} \\
H-optimus-0 & Deformable DETR & 0.7003 {\tiny [0.6663, 0.7258]} & 0.6520 {\tiny [0.6079, 0.6928]} & 0.7564 {\tiny [0.7079, 0.7889]} & 4.344 {\tiny [3.654, 5.047]} \\
\addlinespace
H-optimus-1 & Faster R-CNN & 0.7122 {\tiny [0.6790, 0.7402]} & 0.6557 {\tiny [0.6178, 0.7043]} & \textbf{0.7794} {\tiny [0.7141, 0.8193]} & 4.054 {\tiny [3.238, 4.953]} \\
H-optimus-1 & RetinaNet & 0.7023 {\tiny [0.6665, 0.7294]} & 0.6585 {\tiny [0.6235, 0.7073]} & 0.7524 {\tiny [0.6699, 0.8043]} & 4.310 {\tiny [3.682, 5.044]} \\
H-optimus-1 & Deformable DETR & 0.6554 {\tiny [0.6201, 0.6825]} & 0.6239 {\tiny [0.5758, 0.6716]} & 0.6903 {\tiny [0.6467, 0.7235]} & 3.425 {\tiny [2.699, 4.271]} \\
\addlinespace
ResNet-50 & Faster R-CNN & 0.6905 {\tiny [0.6501, 0.7187]} & 0.6592 {\tiny [0.6141, 0.6965]} & 0.7249 {\tiny [0.6581, 0.7768]} & 4.103 {\tiny [3.549, 4.735]} \\
ResNet-50 & RetinaNet & 0.7193 {\tiny [0.6792, 0.7443]} & 0.6841 {\tiny [0.6372, 0.7214]} & 0.7584 {\tiny [0.6974, 0.7987]} & \textbf{4.694} {\tiny [4.175, 5.251]} \\
ResNet-50 & Deformable DETR & 0.6882 {\tiny [0.6531, 0.7120]} & 0.6546 {\tiny [0.5979, 0.6994]} & 0.7254 {\tiny [0.6778, 0.7668]} & 4.340 {\tiny [3.841, 4.925]} \\
\addlinespace
ResNet-50 (frozen) & Faster R-CNN & 0.6605 {\tiny [0.6203, 0.6963]} & 0.7030 {\tiny [0.6348, 0.7593]} & 0.6228 {\tiny [0.5700, 0.6775]} & 3.774 {\tiny [3.117, 4.596]} \\
ResNet-50 (frozen) & RetinaNet & 0.6693 {\tiny [0.6230, 0.7042]} & 0.7200 {\tiny [0.6664, 0.7655]} & 0.6253 {\tiny [0.5639, 0.6770]} & 4.091 {\tiny [3.513, 4.751]} \\
ResNet-50 (frozen) & Deformable DETR & 0.6925 {\tiny [0.6535, 0.7196]} & 0.7699 {\tiny [0.7306, 0.8018]} & 0.6293 {\tiny [0.5720, 0.6777]} & 4.403 {\tiny [3.909, 4.933]} \\
\bottomrule
\end{tabular}%
}
\end{table}


Table~\ref{tab:midogpp_results} reports detection performance on the test split of MIDOG++ for the 18 frozen backbone-head configurations and two ResNet-50 baselines (fully fine-tuned and frozen). The best configuration for this in-domain use case was the fully fine-tuned ResNet-50 baseline with RetinaNet, which attained the highest F1 (0.7917) in the table. Among the frozen \acp{FM}, the strongest was H-Optimus-0 with RetinaNet (F1 0.7718), which approached but did not match the corresponding fine-tuned baseline.


Performance varied markedly with the choice of detection head. RetinaNet was the most reliable, achieving the top F1 for the ResNet-50 baseline, H-Optimus-0, and Virchow, and never being the worst head for any backbone. In terms of backbone ranking, the two pathology-specific H-Optimus models led among the frozen FMs, closely followed by the Virchow family, with UNI and UNI2-h trailing. The highest recall was obtained by H-Optimus-1 with Faster R-CNN ($0.7859$), though its lower precision ($0.7580$) leaves its F1 ($0.7717$) marginally below the best frozen configuration. H-Optimus-0 with RetinaNet instead achieved the best precision ($0.8197$) among all model configurations, baseline included. The frozen ResNet-50 baseline, which like the \acp{FM} trains only the neck and head, trailed both its fine-tuned counterpart and the strongest frozen FMs (best F1 0.7498 with Deformable DETR), indicating that the competitiveness of the frozen FMs does not arise merely from freezing any ImageNet-pretrained stem but from the pathology-specific representations themselves.

The F1 and FROC-AUC rankings did not always coincide. The clearest case is H-Optimus-1, where Faster R-CNN attained a marginally higher F1 than RetinaNet (0.7717 vs.\ 0.7683) yet a lower FROC-AUC (5.171 vs.\ 5.264); because the ResNet-50 baseline showed the same Faster~R-CNN/RetinaNet FROC ordering, this most likely reflects how each head distributes confidence across operating points rather than a frozen-backbone artifact.

In \ac{OOD} case, the overall performance dropped relative to the in-domain setting (Table~\ref{tab:tupac_results}), and the best F1 was achieved by a frozen \ac{FM} rather than the baseline: H-Optimus-0 with RetinaNet reached the highest F1 ($0.7349$), ahead of the strongest ResNet-50 configuration (RetinaNet, $0.7193$). This reverses the in-domain ordering and indicates that frozen FM features were slightly more robust under domain shift. Virchow with Faster R-CNN attained the best precision ($0.7401$), H-Optimus-1 with Faster R-CNN the best recall ($0.7794$), and the ResNet-50 baseline with RetinaNet the best FROC-AUC ($4.694$). As on MIDOG++, UNI with Faster R-CNN remained the weakest configuration (F1 $0.3952$, FROC-AUC $1.773$), and the relative ordering of backbones was broadly preserved across the two datasets. The frozen ResNet-50 baseline again lagged the frozen \acp{FM} out-of-domain (best F1 0.6925 with Deformable DETR), and unlike the fine-tuned baseline it did not benefit from adaptation to the source domain, reinforcing that the strongest frozen FMs transfer more favourably than a frozen convolutional stem.

\section{Discussion}

Our results give a qualified but clear answer to the central question: the frozen latent spaces of pathology \acp{FM} are rich enough to support dense \ac{MF} detection. As reported above, the strongest frozen configuration (H-Optimus-0 with RetinaNet) comes close to the best end-to-end-trained ResNet-50, and under a Faster R-CNN head the gap effectively vanishes. That backbones trained purely with image-level self-supervision, and kept entirely frozen, can rival a fully fine-tuned convolutional network shows that the spatial information needed to localize \acp{MF} is already present in these representations and recoverable by a lightweight neck and head. A frozen ResNet-50 trained under the identical neck-and-head regime did not reach the same level, which isolates the contribution of the pathology-specific pretraining rather than of the frozen-backbone protocol alone. Performance nonetheless depended substantially on the detection head, an interaction strongest for the weaker backbones. We also observed that Faster R-CNN consistently trailed RetinaNet on FROC-AUC despite competitive F1, a pattern shared by the ResNet-50 baseline and therefore most likely reflecting how each head distributes confidence across operating points rather than a frozen-backbone artifact. Most notably, the in-domain ordering reversed out-of-domain: on TUPAC16 the best F1 came from a frozen FM (H-Optimus-0 with RetinaNet) rather than the baseline, consistent with frozen large-scale features generalizing more gracefully under domain shift.

Several limitations temper these conclusions. The backbones were used strictly frozen, so we characterize the information already present in the pretrained representations, not the ceiling achievable with adaptation. The most natural next step is parameter-efficient fine-tuning, in particular \ac{LoRA}, which has proven effective for \ac{MF} classification with these same \acp{FM}; this would test whether the residual gap to the end-to-end baseline reflects a genuine limit of the frozen representations or merely the absence of light adaptation, though whether the out-of-domain robustness survives such adaptation is itself open. Further limitations include reliance on a single secondary dataset for the OOD evaluation, so the robustness advantage should be read as suggestive, and confidence intervals that are marginal per-configuration estimates, so between-configuration comparisons should account for the overlapping intervals in the tables. 

Overall, our findings support a cautiously optimistic conclusion: the latent spaces of current pathology FMs, despite being shaped entirely by image-level self-supervision, are discriminative and spatially resolved enough to drive \ac{MF} detection competitively with an end-to-end-trained baseline and to transfer slightly more robustly to unseen data, with the most consequential design choices lying in the neck and head--and, prospectively, in lightweight backbone adaptation.

    

\begin{credits}

\ifpeerreview
\subsubsection{\ackname} Withheld for peer review.
\else
\subsubsection{\ackname} M.A. and S.B. acknowledge funding by the Deutsche Forschungsgemeinschaft (DFG, project number: 520330054). C.A.B., V.W., CS, and CK acknowledge funding by the Austrian Research Fund (FWF, project number: I 6555). J.A. acknowledges support by the Bavarian State Ministry of Science and the Arts (project Fokus-TML). K.B. acknowledges funding by the DFG, project number 460333672 CRC1540 EBM. Experiments were supported by the MUSICA HPC cluster at Meduni Vienna and the Julia 2 cluster at Universität Würzburg. Julia 2 hardware is funded by the German Research Foundation (DFG).

\subsubsection{\discintname}
The authors have no competing interests to declare that are
relevant to the content of this article. 

\fi
\end{credits}

\newpage
\begin{acronym}
\acro{FM}[FM]{foundation model}
\acro{HE}[H\&E]{Hematoxylin \& Eosin}
\acro{ROI}[RoI]{region of interest}
\acro{WSI}[WSI]{whole slide image}
\acro{MF}[MF]{mitotic figure}
\acro{NMS}[NMS]{non-maximum suppression}
\acro{OOD}[OOD]{out-of-domain}
\acro{IoU}[IoU]{Intersection over Union}
\acro{AtNorM}[AtNorM]{Atypical and Normal Mitosis}
\acro{AMF}[AMF]{atypical mitotic figure}
\acro{CNN}[CNN]{convolutional neural network}
\acro{ViT}[ViT]{vision transformer}
\acro{LoRA}[LoRA]{Low Rank Adaptation}
\acro{ROC}[ROC]{receiver operating characteristic}
\acro{AUROC}[AUROC]{Area under the Receiver Operating Characteristic Curve}
\end{acronym}

%
%
%
\bibliographystyle{splncs04}
\bibliography{mybibliography}

\end{document}